\documentclass[review]{elsarticle}

\pdfoutput=1
\usepackage{lineno,hyperref}
\usepackage{graphicx}
\usepackage{amsmath}
\usepackage{amssymb}
\usepackage{booktabs}

\begin{document}

\begin{frontmatter}

\title{Towards Simple and Accurate Human Pose Estimation with Stair Network}


\author[firstaddress,secondaddress]{Chenru Jiang}
\ead{chenru.jiang@liverpool.ac.uk}

\author[thirdaddress]{Kaizhu Huang\corref{mycorrespondingauthor}}
\cortext[mycorrespondingauthor]{Corresponding author}

\author[firstaddress]{Shufei Zhang}

\author[secondaddress]{Xinheng Wang}

\author[secondaddress]{Jimin Xiao}

\author[fifthaddress]{Zhenxing Niu}

\author[fourthaddress]{Amir Hussain}

\address[firstaddress]{Department of Computer Science, University of Liverpool, Liverpool L69 7ZX, U.K.}

\address[secondaddress]{Department of Electrical and Electronic Engineering, Xi'an Jiaotong-Liverpool University, Suzhou 215123, China.}

\address[thirdaddress]{Data Science Research Center, Duke Kunshan University, No. 8 Duke Avenue, Kunshan, 215316, China.}

\address[fourthaddress]{School of Computing, Edinburgh Napier University, Edinburgh, EH11 4BN, UK.}

\address[fifthaddress]{School of  Electronic Engineering, Xidian University, Xian, 710000, China.}




\begin{abstract}
   In this paper, we focus on tackling the precise keypoint coordinates regression task. Most existing approaches adopt complicated networks with a large number of parameters, leading to a heavy model with poor cost-effectiveness in practice. To overcome this limitation, we develop a small yet discrimicative model called STair Network, which can be simply stacked towards an accurate multi-stage pose estimation system. Specifically, to reduce computational cost, STair Network is composed of novel basic feature extraction blocks which focus on promoting feature diversity and obtaining rich local representations with fewer parameters, enabling a satisfactory balance on efficiency and performance. To further improve the performance, we introduce two mechanisms with negligible computational cost, focusing on feature fusion and replenish. We demonstrate the effectiveness of the STair Network on two standard datasets, e.g., 1-stage STair Network achieves a higher accuracy than HRNet by 5.5\% on COCO test dataset with 80\% fewer parameters and 68\% fewer GFLOPs. \footnote{The code is released at https://github.com/ssr0512} \footnote{© 2022 IEEE.  Personal use of this material is permitted.  Permission from IEEE must be obtained for all other uses, in any current or future media, including reprinting/republishing this material for advertising or promotional purposes, creating new collective works, for resale or redistribution to servers or lists, or reuse of any copyrighted component of this work in other works}
\end{abstract}

\begin{keyword}
Stair Network, Human Pose Estimation, Feature Diversity
\end{keyword}

\end{frontmatter}


\section{Introduction}
\label{sec:intro}

Human pose estimation is one fundamental yet challenging task to estimate precise human joint coordinates (eyes, ears, shoulders, elbows, wrists, knees, etc.). It is essential for various high-level visual understanding tasks such as human action recognition~\cite{Wang_2013_CVPR, liang2014expressive, zhu2019action, wang2014robust}, video surveillance~\cite{li2019state}, and tracking~\cite{xiaohan2015joint, cho2013adaptive}. In recent years, there have been significant advances from single pose estimation~\cite{tompson2014joint, toshev2014deeppose, wei2016convolutional, andriluka2009pictorial, newell2016stacked, yang2017learning, witoonchart2017application} to multiple pose estimation~\cite{cao2017realtime, chen2018cascaded, insafutdinov2016deepercut, sun2019deep, newell2017associative, pishchulin2016deepcut}. These methods can be categorized into bottom-up~\cite{cao2017realtime, insafutdinov2016deepercut, newell2017associative, pishchulin2016deepcut, papandreou2018personlab, cheng2020higherhrnet} and top-down~\cite{chen2018cascaded, he2017mask, huang2020devil, papandreou2017towards, sun2019deep, xiao2018simple, zhang2020distribution} methods. Top-down methods are gaining more popularity due to their higher accuracy.

\begin{figure}[htbp]
  \centering 
  \includegraphics[width=6.5cm]{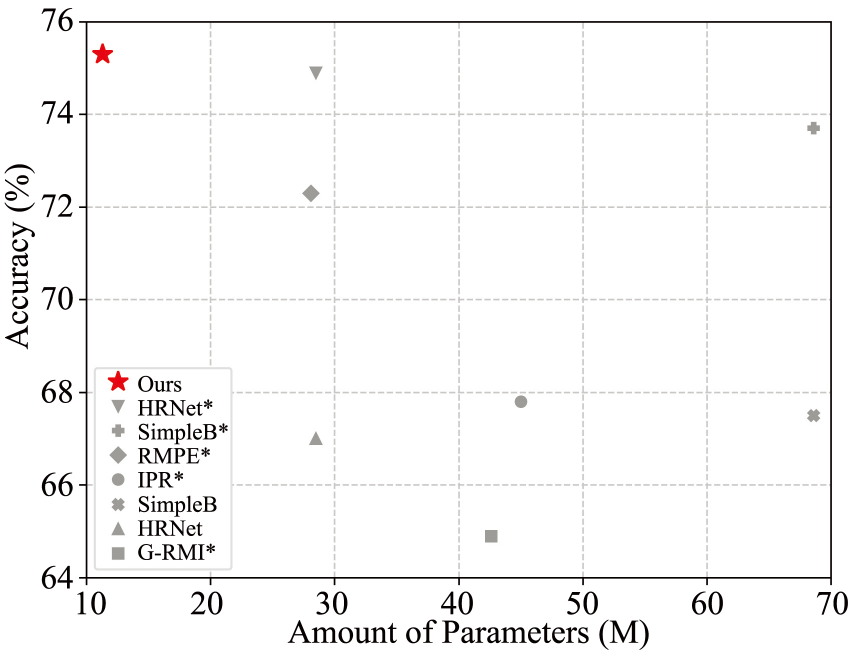}
  \caption{Performance vs. parameter number of various methods on COCO test data with 384$\times$288 input size. Method in red is 3-stage STair Network, and * means the method adopts pre-training.} 
  \label{fig:STNet_P}
\end{figure}

\begin{figure*}
  \centering 
  \includegraphics[width=12cm]{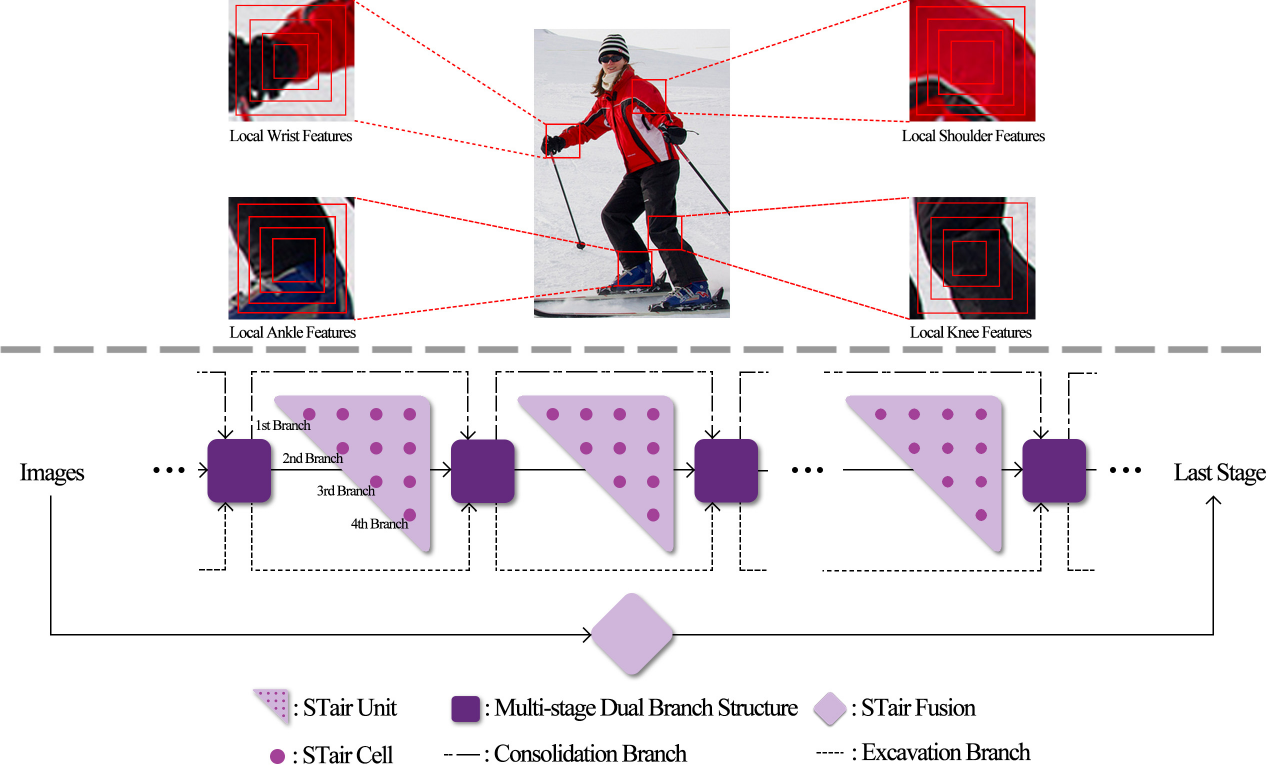}
  \caption{The upper figure part illustrates the local features of multi-scale keypoints, where the four red boxes on local features are the multiple receptive fields of STair Cell. The lower figure part demonstrates the STair Network structure.} 
  \label{fig:STNet}
\end{figure*}

In existing top-down methods, multi-stage structures~\cite{zhang2019human, yang2017learning, zhang2019fast} and knowledge distillation~\cite{zhang2019fast, li2021online} are commonly adopted for pose estimation. However, two main drawbacks limit both the efficiency and the accuracy. One defect is that single receptive field for local features is commonly adopted in existing methods. Due to the insufficient local features extraction, current network structures are usually deep and complicated for good performance. Fig.~\ref{fig:STNet_P} shows a number of existing popular methods which commonly adopt a heavy model with a large number of parameters. Although multi-scale module structure~\cite{lin2017feature, newell2016stacked} is commonly 
designed to aggregate information, the feature diversity is still coarse and insufficient for regression tasks. The top part of Fig.~\ref{fig:STNet} illustrates a scenario that the single receptive field (the smallest red rectangles) is deficient to distinguish background or different torsos. For knowledge distillation, complicated teacher networks and iterative training process are indispensable. The other defect is that the information loss is inevitable in these structures. We observe that, for localization tasks, the position error will be accumulated and enlarged after the iterative down/upsampling process. In addition, these methods seldom consider the important high-frequency texture representations~\cite{wang2020high}.

\begin{figure}[htbp]
    \centering
    \includegraphics[width=9cm]{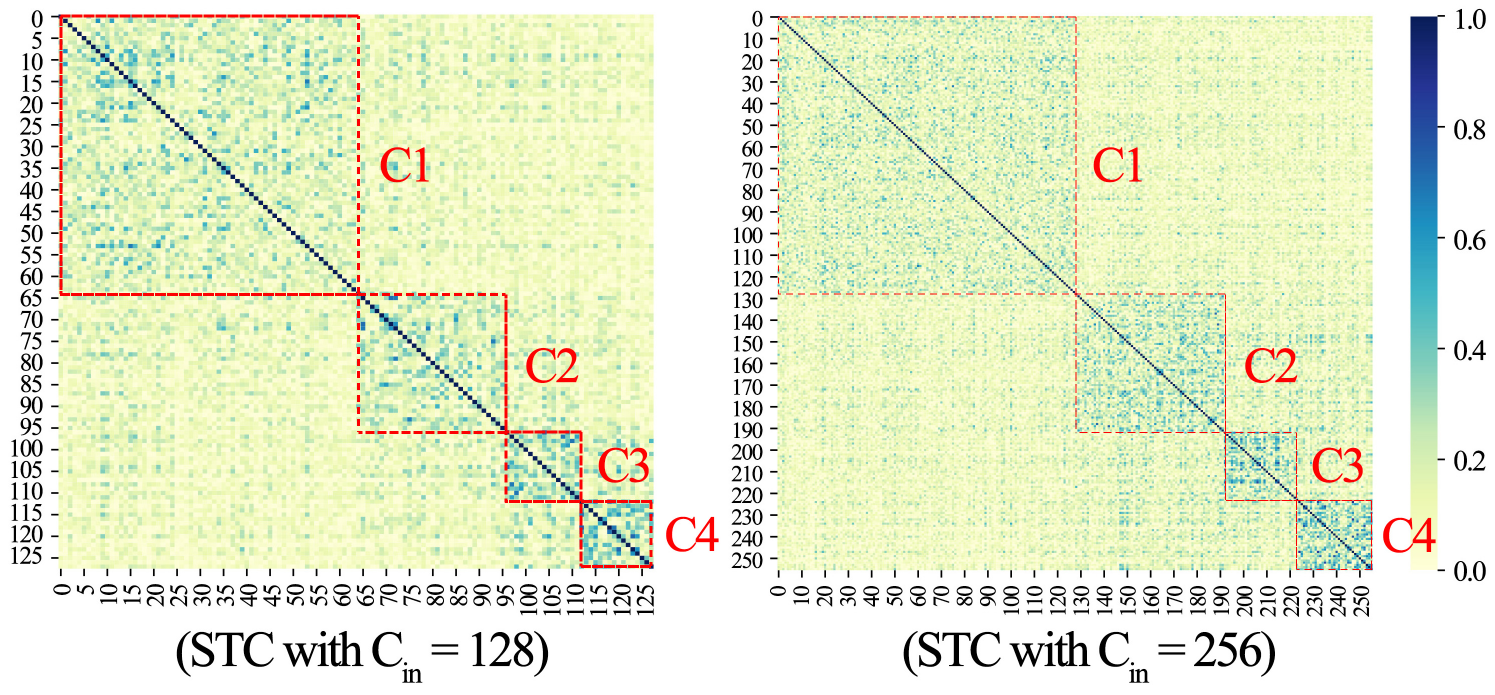}
    \caption{Branch correlation of STC. Deeper color means stronger correlation. C1, C2, C3 and C4 are corresponding to the four branches of STC in Fig. 4. Better viewed in color.}
    \label{fig:channel_Diff}
\end{figure}

To reduce computational cost and still achieve superior performance, we design a novel basic feature extraction block, STair Cell (STC), to simultaneously pursue advantageous feature diversity and high efficiency. In each block, multiple receptive fields structure is introduced to promote local feature diversity and aggregate rich local representations, enabling the block to obtain stronger discriminative capability on multi-scale keypoints or background (as illustrated multiple red rectangles in the top part of Fig.~\ref{fig:STNet}). Meanwhile, a lightweight context attention is embedded to enlarge the features discrimination. Fig.~\ref{fig:channel_Diff} visualizes the correlations within STC where multiple receptive fields focus on aggregating different local features and bring rich information difference. For efficiency, the block channel number is gradually halved and depthwise separable mechanism is adopted to attain lower computational cost.

To reduce computational cost and still achieve superior performance, we design a novel basic feature extraction block, STair Cell (STC), to simultaneously pursue advantageous feature diversity and high efficiency. In each block, multiple receptive fields structure is introduced to promote local feature diversity and aggregate rich local representations, enabling the block to obtain stronger discriminative capability on multi-scale keypoints or background (as illustrated multiple red rectangles in the top part of Fig.~\ref{fig:STNet}). Meanwhile, a lightweight context attention is embedded to enlarge the features discrimination. Fig.~\ref{fig:channel_Diff} visualizes the correlations within STC where multiple receptive fields focus on aggregating different local features and bring rich information difference. For efficiency, the block channel number is gradually halved and depthwise separable mechanism is adopted to attain lower computational cost.

We encapsulate STCs to build a simple multi-branch module termed STair Unit (STU), which is illustrated as triangles in Fig.~\ref{fig:STNet}. We propose to tackle the information loss problem from two aspects. Within each unit, down/upsampling is not taken in the first branch such that the network can consistently keep the high resolution feature maps to alleviate information loss. Outside each unit, we design a lightweight structure to focus on feature re-usage and re-exploitation among multiple units. Illustrated as squares in Fig.~\ref{fig:STNet}, such structure can be readily attached after each unit for enhancing feature utilization. Meanwhile, we try to reserve high-frequency texture representations (illustrated as diamonds in Fig.~\ref{fig:STNet}), to supply these important features at the last unit for precise localization.

STair Units can be simply stacked to form a multi-stage network for the coarse-to-fine estimation. Significantly different from knowledge distillation, by enhancing the feature extraction block while alleviating information loss problem, the large teacher network is not needed in our network. In a nutshell, our contributions are three-fold:

\begin{enumerate}
    \item To attain a low computational cost and accuracy framework, we propose a lightweight yet effective basic feature extraction block to focus on extracting more diverse local representations, enabling  better capability to localize different keypoints with less parameters.

    \item To reduce information loss during the training, the first branch within each STC consistently keeps high resolution feature maps. Meanwhile, two efficient mechanisms are proposed to connect STCs to enhance feature utilization and replenish.

    \item STNet attains new SOTA performance  with lower computational cost on standard benchmarks. Specifically, even 1-stage network can boost the accuracy by 5.5\% over previous SOTA network~\cite{sun2019deep} with only 20\% parameters when evaluated on COCO test datasets. 
\end{enumerate}

\section{Related Work}
\paragraph{Feature Extraction Unit} Residual block is a popular basic feature extraction unit and commonly adopted in existing pose estimation methods~\cite{he2016deep, sandler2018mobilenetv2, newell2016stacked, chen2018cascaded, xiao2018simple, sun2019deep}. By the efficient bottleneck design with the skip connection, the residual block can be utilized to form much deeper and more complicated structures. Meanwhile, some novel methods can also assist to improve the residual block capability. For instance, SENet~\cite{hu2018squeeze} embeds channel attention after each block to reweight each feature map.
SKNet~\cite{li2019selective} increases a convolution with the different kernel size in the block, and adopts soft attention to select useful channels. ResNeXt~\cite{xie2017aggregated} applies group convolution~\cite{krizhevsky2012imagenet} in order to obtain rich features from different subspace. Currently, RSN~\cite{cai2020learning} proposes to equally split channels into 4 branches in one block, and increases the convolutions and inner connections to
fuse features gradually. However, RSN relies heavily on complicated and advanced platforms MegDL, and the performance RSN\_18 drops significantly (3.2\%) when Pytorch is adopted. As demonstrated in Sec.~\ref{STC}, the basic feature extraction blocks of this paper consist of four sequential convolutions, where the channel dimension is halved and receptive field is doubled gradually to capture rich diverse local features while maintaining high efficiency.

\paragraph{Multi-stage Learning Structure} Multi-stage structure proves suitable to attain accurate localization, which has been widely used in various recent approaches~\cite{li2019rethinking, sun2019deep, zhang2019human, yang2017learning} for refining predictions gradually. Hourglass~\cite{newell2016stacked} and fast pose~\cite{zhang2019fast} take the same sub-module to form multi-stage networks. MSPN~\cite{li2019rethinking} follows the hourglass design yet doubles the channel dimension gradually after each downsampling. Among multiple stages, existing methods~\cite{chu2017multi, ke2018multi, zhang2019fast} focus on feature attention, and intermediate supervision~\cite{wei2016convolutional} to adjust and supervise the sub-module learning process. Differently, as shown in Sec.~\ref{MDBS}, we propose a multi-stage dual branch structure which mainly focus on feature re-usage and re-exploitation among stages.

\paragraph{Low-frequency Structure Feature Fusion} In the traditional methods~\cite{cai2020learning, chen2018cascaded, sun2019deep}, multi-scale feature fusion is adopted commonly to extract low-frequency semantic information. Hourglass network~\cite{newell2016stacked} proposes a U-shape structure to attain multi-scale feature fusion within each stage. Later works such as cascaded pyramid network~\cite{chen2018cascaded} execute a multi-scale fusion process between high-to-low and low-to-high structures. HRNet~\cite{sun2019deep} sets up four parallel branches and aggregates branch features iteratively. Intuitively, current methods all work on low-frequency structure feature fusion. However, high-frequency information is another critical factor~\cite{wang2020high} for the precise localization tasks as it provides rich texture representations. Unfortunately, high-frequency information is hardly reserved when networks go deeper. Contrary to conventional approaches, in Sec.~\ref{STair Fusion} we develop an effective method to directly replenish abundant multi-scale high-frequency representations.

\section{Our Approach}

In this section, we  first detail our basic component  STair Cell. After that, we introduce the multi-scale STair Unit structure consisting of a number of STCs that maintains high resolution feature maps from end to end. Next, we will describe the Multi-stage Dual Branch Structure and STair Fusion mechanism to further enhance the network.

\subsection{STair Cell}
\label{STC}

\paragraph{STair Convolution} For each STair Cell, we propose a sequential stair-shape atrous convolution structure to aggregate significant diverse local features for different scale keypoints through multiple receptive field sizes. In contrast to traditional convolutions, atrous convolution~\cite{chen2017deeplab} helps to attain larger receptive fields without increasing the computational cost. The kernel sizes of atrous convolution layers are calculated by 
$K = k + (k-1)\times(R-1)$, where $K$ is the equivalent kernel size, $k$ is the practical kernel size, and $R$ is the dilation rate.

\begin{figure}[htbp]
  \centering
  \includegraphics[width=8cm]{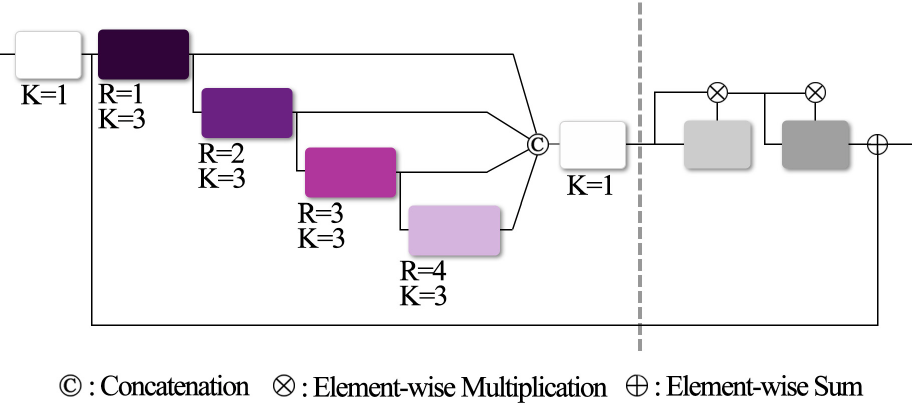}
  \caption{STair Cell structure. Purple rectangles indicate atrous convolution processes with multiple dilation rates, gray rectangles describe the embedded context attentions.}
  \label{fig:STC}
\end{figure}

As shown in Fig.~\ref{fig:STC}, one stair shape convolution structure contains four atrous convolution layers with the different dilation rates ($R = 1, 2, 3, 4$). We notice that the atrous convolution has an inherent gridding effect which leads to information inconsistency~\cite{fang2020face}. To alleviate such gridding effect, we keep the dilation rate as $1$ in the top base convolution layer, make it as the same as the normal convolution.  In STC, as the receptive field size is increased, the channel number is reduced in half for the efficiency purpose. Concretely, the total computational cost of four-branch structure are: $4T (T= k \times k \times C_{in} \times C_{out} \times H \times W$), where the computational cost depends multiplicatively on the number of input channels $C_{in}$, the number of output channels $C_{out}$, the kernel size $k$, and the feature map size $H \times W$. The channel halving strategy factorizes conventional mechanism into four parts and the amount of parameters are:
\begin{align}
    \frac{1}{2}T + \frac{1}{8}T + \frac{1}{32}T + \frac{1}{64}T = \frac{43}{64}T
    \label{halve}
\end{align}

Thus, the channel halving strategy attains $4/(\frac{43}{64})\approx6$ times computation reduction than the conventional structure. Table~\ref{table:channel_halving} demonstrates the computational cost comparison of channel halving strategy. More channels might provide richer diverse features but hinder structure efficiency, which is not pursued in this work. Particularly, we have investigated various structures for STC which are shown in ablation parts. However, the current structure demonstrates the best performance for aggregating diverse local features with high efficiency.

\setlength{\tabcolsep}{2pt}
\begin{table}[htbp]
\footnotesize
\caption{Results of STC with/without channel halving strategy.}
\begin{center}
\label{table:channel_halving}
\begin{tabular}{ccccccc}
\hline
\noalign{\smallskip}
Method & pretrain & Input Size & Halve & \#Params & GFLOPs & AP\\
\noalign{\smallskip}
\hline
\noalign{\smallskip} 
\hline
\noalign{\smallskip} 
1-stage & N & 256 $\times$ 192 & $\checkmark$ & 5.7M & 2.3 & 72.1\\
1-stage & N & 256 $\times$ 192 & $\times$ & 8.7M & 3.0 & 73.0\\
\noalign{\smallskip} 
\hline
\end{tabular}
\end{center}
\end{table}

Table~\ref{table:STC&STNet} lists the detail parameter configurations of STC and STU. Since the inputs of each atrous convolution are based on the previous layers, the receptive fields are gradually superimposed. Thus, the fourth branch of the STC contains four receptive fields equivalently. In this regard, STC enables to extract rich diverse local representations for the network. As demonstrated in Fig.~\ref{fig:channel_Diff}, the correlations between different STC branches ($Ci$ and $Cj$, $i \neq j$) are low, meaning that multiple branches with various receptive fields focus on aggregating different local features. Consequently, four-branch STair Units can obtain more dense diverse features with STCs, as shown in the second part of Table~\ref{table:STC&STNet}.

\setlength{\tabcolsep}{2pt}
\begin{table}[htbp] 
\footnotesize
\caption{Parameter Configuration of STair Cells and STair Units. M means that the feature map size is maintained, and H means the feature map size is halved by downsampling.}
\begin{center}
\label{table:STC&STNet}
\begin{tabular}{cccc}
\hline
\noalign{\smallskip}
STC Branch Index & Feature Map Size & Channel Number & Kernel Size  \\
\noalign{\smallskip}
\hline
\noalign{\smallskip}
\hline
\noalign{\smallskip}
c = 1 & M & 16 & 3 \\
c = 2 & M & 8 & 3, 5  \\
c = 3 & M & 4 & 3, 5, 7 \\
c = 4 & M & 4 & 3, 5, 7, 9 \\
\noalign{\smallskip} 
\hline
\noalign{\smallskip}
STU Branch Index & Feature Map Size & Channel Number & Kernel Size  \\
\noalign{\smallskip}
\hline
\noalign{\smallskip}
\hline
\noalign{\smallskip}
b = 1 & M & 32 & 3, 5, 7, 9 \\
b = 2 & H & 64 & 5, 7, 9, 11  \\
b = 3 & H & 128 & 7, 9, 11, 13 \\
b = 4 & H & 256 & 9, 11, 13, 15 \\
\noalign{\smallskip} 
\hline
\end{tabular}
\end{center}
\end{table}

To further reduce the parameters, we adapt the depthwise separable convolution~\cite{howard2017mobilenets} to our STC, which factorizes a standard convolution into a depthwise operation and a pointwise operation. First, each atrous convolution of STC exploits depthwise operation to apply a single filter for each input channel. Then, the pointwise operation applies a $1\times1$ convolution to combine the depthwise outputs. There is a significant difference on the amounts of network parameters between traditional convolution and depthwise separable convolution. Specifically, the total computational cost of a standard convolution are: $T$, The depthwise separable convolution splits standard convolution operation into two parts: $ T / k^{2} + T / C_{out}$. After decomposing convolution as a two-step process of filtering and combining, we can achieve the parameter reduction as below:
\begin{align}
    &\frac{T/k^{2} + T/C_{out}}{T} = \frac{1}{k^{2}} + \frac{1}{C_{out}}
    \label{funDepth}
\end{align}

After the stair-shape convolution, we concatenate four level features and pass them to the lightweight context attention part. 

\paragraph{Mix Attention} For the postprocessing of STC, we consider separately for different local features so that the block attains a stronger feature diversity and fine-grained recalibration. The proposed mix attention strategy is shown in the right part of Fig.~\ref{fig:STC} which aims to reinforce the extracted diverse features through both channel and spatial dimensions. For calculation efficiency, we introduce soft reduction rate to the channel dimension attention, which is dynamically enlarged as the channels of STU is increased. For spatial dimension attention, STC applies average and max pooling methods to generate only two masks for keeping low computational cost. Table~\ref{table:mix_attention} demonstrates that the mechanism is effective to enhance feature diversity with negligible computational cost. In addition, we maintain skip connection to support gradients propagation of the network.

\setlength{\tabcolsep}{2pt}
\begin{table}[htbp]
\footnotesize
\caption{Comparison of STC with/without mix attention.}
\begin{center}
\label{table:mix_attention}
\begin{tabular}{ccccccc}
\hline
\noalign{\smallskip}
Method & pretrain & Input Size & \#Params & GFLOPs & Mix Attention & AP\\
\noalign{\smallskip}
\hline
\noalign{\smallskip} 
\hline
\noalign{\smallskip} 
1-stage & N & 256 $\times$ 192 & 5.6M & 2.3 & $\times$ & 71.5\\
1-stage & N & 256 $\times$ 192 & 5.7M & 2.3 & $\checkmark$ & 72.1\\
\noalign{\smallskip} 
\hline
\end{tabular}
\end{center}
\end{table}

\subsection{STair Unit}
\begin{figure}[htbp] 
  \centering
  \includegraphics[height=5cm]{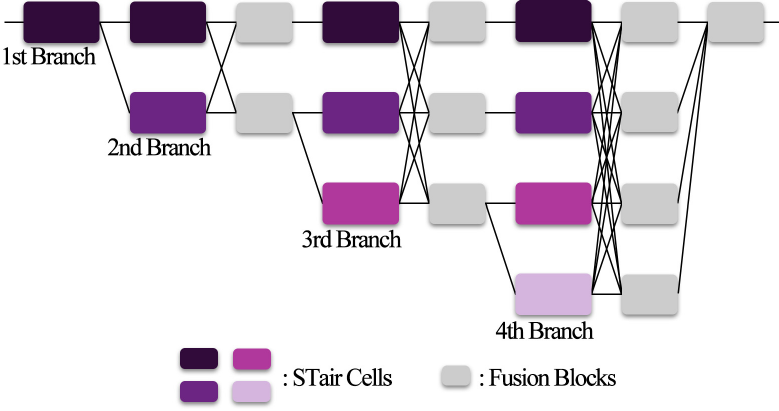}
  \caption{Structure of each STair Unit. Purple and gray rectangles describe respectively STair Cells and multi-scale feature fusion process.}
  \label{fig:STU}
\end{figure}

STU can be stacked to compose a multi-stage learning mechanism as demonstrated in Fig.~\ref{fig:STNet}. Inspired by HRNet~\cite{sun2019deep}, each STU adopts a multi-branch structure to attain multi-scale feature fusion. Differently, with the superior STCs, the basic extraction block number of four branches in each STU is: (4,3,2,1), whilst that of the HRNet is: 4x(4,3,2,1). Thus, the computational cost is much less than HRNet. Meanwhile, as seen in the bottom part of Table~\ref{table:STC&STNet},  we list the detail kernel size range of the units. More receptive field of STU (3-15) can obtain rich representations than HRNet (3-9). The inner structure of the units is illustrated in Fig.~\ref{fig:STU} where the purple rectangles describe STCs and the gray rectangles are the multi-scale feature fusion process. To reduce information loss, the sizes of feature maps are reduced in half but branch channels are doubled gradually from top to bottom, and the top branch consistently maintains high resolution featrue map size from end to end. To make the network to obtain rich multi-scale features for precise predictions, we design the four-stage feature fusion in one unit. In feature fusion process, each sub-branch iteratively aggregates the features which are downsampled or upsampled from other parallel sub-branches. 

\subsection{Multi-stage Dual Branch Structure}
\label{MDBS}
For feature re-usage, skip connections are widely applied between network stages~\cite{chen2018cascaded, sandler2018mobilenetv2}. However, existing methods pay little attention to feature re-exploration, since the parameters grow quadratically as the densely connected path width increases linearly. In this section, we will explain the limitations of ResNet~\cite{he2016deep} which focus on feature re-usage, and DenseNet~\cite{huang2017densely} which focus on feature re-exploitation first. We then present the Multi-stage Dual Branch Structure (MDBS), which enjoys the benefits from both path topologies for learning good representations.

\paragraph{Revisiting ResNet and DenseNet} Traditional convolution feed-forward networks utilize the output of $L^{th}$ layer as the input for (${L+1}$)$^{th}$ layer, which can be summarized as: $X_{L} = H_{L}$($X_{L}-1$). ($X_{L}$ is the output of $L^{th}$ layer, and $H_{L}$ are the composite operations such as batch normalization, rectified linear unit, or convolution). ResNet proposes a skip connection which bypasses the non-linear transformations with an identity function:
\begin{align}
    X_{L} = H_{L}(X_{L-1}) + X_{L-1}.
    \label{funRes}
\end{align}

The main contribution of ResNet is that the gradients can propagate directly through the identity function to relieve gradient attenuation and reinforce the feature re-usage in deeper layers. However, ResNet neglects the features re-exploration and the summation process in Eq.~(\ref{funRes}) may impede the information flow in the network. After that, Huang et al.~\cite{huang2017densely} propose DenseNet, where the skip connections are used to concatenate the inputs to the outputs instead of adding operation in order to improve the information flow between layers. Consequently, the $L_{th}$ layer will receive the feature maps of all preceding layers as inputs:
\begin{align}
    X_{L} = H_{L}(X_{0}||X_{1}||...|| X_{L-1}).
    \label{funDense}
\end{align}

$X_{0}||X_{1}||. . .||X_{L-1}$ refers to the feature maps concatenation ($||$) of layer $0, …, (L-1)^{th}$. Since the width of the densely connected path and the cost of GPU memory linearly increase as the network goes deeper, building a deeper and wider densenet is substantially restricted.

\paragraph{Multi-stage Dual Branch Structure} MDBS combines advantages of both path learning~\cite{he2016deep, huang2017densely, chen2017dual}, able to reinforce information re-usage and re-exploration among multiple STair Units. In each stage of MDBS, we feed the unit outputs  into a simple 1$\times$1 conv to generate two branches: consolidation branch and excavation branch. Similar to DenseNet, we set a constant $n$ as the growth rate in excavation branch, which is half of the unit first branch channel number. The small number of $n$ helps slow the increase on width of branch excavation and the GPU memory occupation. The stage inputs, consolidation branch and excavation branch of MDBS can be expressed as follows:
\begin{gather}
    C_{i} = H_{i}(X_{i}^{out})[0:n] + H_{i-1}(X_{i-1}^{out})[0:n],
    \label{funDualRes}\\
    E_{i} = H_{i}(X_{i}^{out})[n:end]||...||H_{0}(X_{0}^{out})[n:end],
    \label{funDualDense}\\
    X_{i+1}^{in} = C_{i}||E_{i},
    \label{funDualInput}
\end{gather}

$X_{i}^{out}$ are the MDBS outputs of stage $i$ and $X_{i+1}^{in}$ are the STU inputs of the stage $i+1$. $H_{i}$ is the shared composite operations that the outputs are equally separated into two parts ($[0:n], [n:end]$) for consolidation and excavation branches. As exhibited in Eq.~(\ref{funDualRes}) and Eq.~(\ref{funDualDense}), element-wise summation is applied on consolidation branch for features re-usage, and concatenation operation is applied on excavation branch for the features re-exploration. After that, as listed in Eq.~(\ref{funDualInput}), two branch features are concatenated together as the inputs for the next STU. As shown in in Fig.~\ref{fig:STNet}, the method can be readily attached after each unit for enhancing feature utilization.

\subsection{STair Fusion}
\label{STair Fusion}
\begin{figure}[htbp]
  \centering
  \includegraphics[height=5cm]{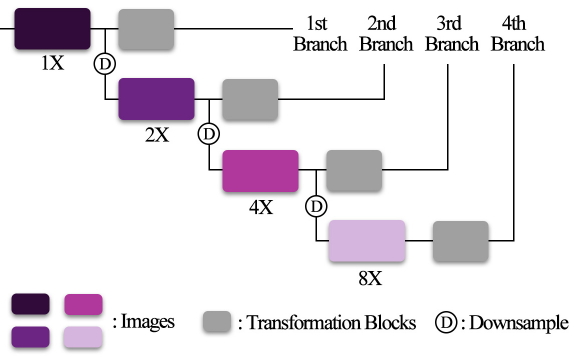}
  \caption{STair Fusion structure. Purple rectangles are multi-scale original images, and gray rectangles are transformation blocks for matching channel with STU multiple branches.}
  \label{fig:SFF}
\end{figure}

In the existing methods~\cite{cai2020learning, chen2018cascaded, sun2019deep}, multi-scale structures are adopted to focus on low-frequency semantic feature fusion. However, the quantity of high-frequency features is another critical factor for precise localization tasks~\cite{wang2020high} as the features contain rich texture representations and better discrimination ability. To this end, we present a STair Fusion (STF) mechanism to replenish multi-scale high-frequency features to the network. As shown in Fig.~\ref{fig:SFF}, STF adopts downsampling to generate multi-scale images. After that, we apply lightweight transformation blocks at each scale to change channels for matching multiple STU branches. The transformation blocks contain just four layers which help to reserve more high-frequency texture features. The reserved high-frequency representations are then concatenated with multi-scale low-frequency features of STU respectively. For the fusion stage, we take upsampling and deconvolution methods to gradually enlarge small-scale concatenated feature maps. We then sum four-scale feature maps to attain multi-scale fusion. For efficiency purpose, we apply a reduction factors on the transformation blocks, which is similar to STC context attention. Significantly, we attempt to modify this mechanism into a multi-stage design. The multi-stage design combines high-frequency features with low-frequency features after each STU, where the high-frequency information may still be lost in the next unit, and the results of single STF is better. Therefore, we adopt the single fusion at the end of the network.

\section{Experiments}
STair Network is evaluated on two standard human pose estimation datasets, i.e. the  COCO keypoint-2017 dataset~\cite{lin2014microsoft} and MPII keypoint~\cite{andriluka20142d} dataset. We follow the process of~\cite{huang2020devil} to decode the predicted heatmaps. The Adam optimizer~\cite{kingma2014adam} is used in training, and we develop STNet on 4 NVIDIA 2080 Ti GPUs.

\paragraph{Results on COCO} This dataset contains over 200,000 images and 250,000 human instances labeled with 17 keypoints. The evaluation metric of the dataset is based on \textit{Object Keypoint Similarity (OKS)}. Standard Average Precision (AP)  and Average Recall (AR)  are used as the metric. AP and AR denotes the mean value of 10 OKS numbers (OKS = 0.50, 0.55, … , 0.95),  AP$^{50}$ denotes AP at OKS = 0.5,  AP$^{M}$ denotes the AP for medium objects. For this dataset, we follow HRNet and adopt the same human instance detector to provide human bounding boxes for both the validation and  test sets. Table~\ref{table:cocoALL_T} lists the results of STNet and other competitive methods on the COCO test-dev dataset.The trunk branch channel number of STNet is 32.

\setlength{\tabcolsep}{1pt}
\begin{table*}[h!]
\tiny
\caption{Comparison results on the COCO test-dev dataset. \#Params are total parameters of the networks, and GFLOPs are total computational cost of the methods. * means STC without channel halving strategy.}
\label{table:cocoALL_T}
\begin{center}
\begin{tabular}{llcccccccccc}
\hline
\noalign{\smallskip}
Method & Backbone & Input Size & Pretrain & \#Params & GFLOPs & AP & AP$^{50}$ & AP$^{75}$ & AP$^{M}$ & AP$^{L}$ & AR \\ 
\noalign{\smallskip} 
\hline
\noalign{\smallskip} 
\hline
\noalign{\smallskip} 
Mask-RCNN~\cite{he2017mask} & ResNet\_50\_FPN & - & Y & - & - & 63.1 & 87.3 & 68.7 & 57.8 & 71.4 & - \\
G-RMI~\cite{papandreou2017towards} & ResNet\_101 & 353$\times$257 & Y & 42.6M & 57.0 & 64.9 & 85.5 & 71.3 & 62.3 & 70.0 & 69.7 \\

HRNet & HRNet\_w32 & 384$\times$288 & N & 28.5M & 16.0 & 67.0 & 85.4 & 74.3 & 64.8 & 73.4 & 78.1\\

IPR~\cite{sun2018integral} & ResNet\_101 & 256$\times$256 & Y & 45.0M & 11.0 & 67.8 & 88.2 & 74.8 & 63.9 & 74.0 & - \\
G-RMI + extra data~\cite{papandreou2017towards} & ResNet\_101 & 353$\times$257 & Y & 42.6M & 57.0 & 68.5 & 87.1 & 75.5 & 65.8 & 73.3 & 73.3 \\
SimpleBaseline~\cite{xiao2018simple} & ResNet\_152 & 384$\times$288 & N & 68.6M & 35.5 & 69.5 & 90.1 & 77.0 & 66.4 & 75.3 & 75.5\\
SimpleBaseline & ResNet\_50 & 384$\times$288 & N & 34.0M & 20.2 & 70.4 & 90.7 & 77.5 & 66.7 & 76.9 & 75.8\\
SimpleBaseline & ResNet\_101 & 384$\times$288 & N & 53.0M & 27.9 & 71.9 & 91.1 & 79.8 & 68.7 & 78.0 & 77.4\\
CPN~\cite{chen2018cascaded} & Resnet\_Inception & 384$\times$288 & Y & - & - & 72.1 & 91.4 & 80.0 & 68.7 & 77.2 & 78.5 \\
RMPE~\cite{fang2017rmpe} & PyraNet~\cite{yang2017learning} & 320$\times$256 & - & 28.1M & 26.7 & 72.3 & 89.2 & 79.1 & 68.0 & 78.6 & - \\
CFN~\cite{huang2017coarse} & \multicolumn{1}{c}{-} & - & Y & - & - & 72.6 & 86.1 & 69.7 & 78.3 & 64.1 & - \\
CPN (ensemble) & ResNet\_Inception & 384$\times$288 & Y & - & - & 73.0 & 91.7 & 80.9 & 69.5 & 78.1 & 79.0 \\
SimpleBaseline & ResNet\_152 & 384$\times$288 & Y & 68.6M & 35.6 & 73.7 & 91.9 & 81.1 & 70.3 & 80.0 & 79.0 \\ 
HRNet & HRNet\_w32 & 384$\times$288 & Y & 28.5M & 16.0 & 74.9 & 92.5 & 82.8 & 71.3 & 80.9 & 80.1\\
HRNet & HRNet\_w48 & 384$\times$288 & Y & 63.6M & 32.9 & 75.5 & 92.5 & 83.3 & 71.9 & 81.5 & 80.5\\
TokenPose~\cite{li2021tokenpose} & L/D24 & 384$\times$288 & Y & 29.8M & 22.1 & 75.9 & 92.3 & 83.4 & 72.2 & 82.1 & 80.8\\
\noalign{\smallskip} 
\hline
\noalign{\smallskip} 
1-stage & STNet & 256$\times$192 & N & 5.7M & 2.3 & 71.4 & 91.0 & 79.0 & 68.1 & 77.2 & 76.8 \\
1-stage & STNet & 384$\times$288 & N & 5.7M & 5.2 & 72.5 & 91.0 & 79.7 & 68.9 & 78.5 & 77.8 \\

2-stage & STNet & 256$\times$192 & N & 8.5M & 2.9 & 73.4 & 91.7 & 81.1 & 70.2 & 78.9 & 78.7 \\
2-stage & STNet & 384$\times$288 & N & 8.5M & 6.5 & 74.8 & 92.0 & 82.1 & 71.4 & 80.7 & 80.0 \\

3-stage & STNet & 256$\times$192 & N & 11.3M & 3.5 & 74.1 & 91.8 & 81.8 & 71.0 & 79.7 & 79.4 \\
3-stage & STNet & 384$\times$288 & N & 11.3M & 7.8 & 75.3 & 92.1 & 82.7 & 71.8 & 81.2 & 80.4 \\
3-stage* & STNet & 384$\times$288 & N & 20.3M & 12.6 & 75.9 & 92.3 & 83.4 & 72.5 & 81.8 & 81.1 \\
\noalign{\smallskip} 
\hline
\end{tabular}
\end{center}
\end{table*}

As observed,  without pre-training, our proposed method achieves very encouraging 75.3\% AP score on the 3-stage backbone with the 384$\times$288 input size. This result is higher than all the comparison models that are even pre-trained. Specifically, compared with HRNet\_32, the parameters of STNet is fewer over 60\% (11.3M), and the GFLOPs is less over 50\% (7.8). Compared with ResNet\_152 backbone of SimpleBaseline, the gain of 3-stage network is 1.6\% with 384$\times$288 input size, and the parameters of STNet is over 83\% fewer, and the reduction on GFLOPs is nearly 80\%. It is also noted that, the computational cost of STNet is the least among these popular methods. In contrast to the existing methods, the least reduction on parameters and GFLOPs achieved by  3-stage also attains 60\% and 30\% respectively. Significantly, with the same 384$\times$288 input size, 1-stage STNet obtains 5.5\% improvements with 80\% drop on parameters and 68\% drop in GFLOPs when compared with no pretrained HRNet\_32. In addition, STNet can achieve higher accuracy (75.9\%) once channel halving strategy is abandoned. We further report the comparisons results on the COCO validation set in Table~\ref{table:cocoALL_V}.

\setlength{\tabcolsep}{2pt}
\begin{table*}[h!]
\tiny
\caption{Comparison results on the COCO validation set. OHKM means online hard keypoints mining~\cite{chen2018cascaded}. Pretain means the method is/isn't pretrained on the ImageNet classification task. * means STC without channel halving strategy.}
\begin{center}
\label{table:cocoALL_V}
\begin{tabular}{llcccccccccc}
\hline\noalign{\smallskip}
Method & Backbone & Input Size & Pretrain & \#Params & GFLOPs & AP & AP$^{50}$ & AP$^{75}$ & AP$^{M}$ & AP$^{L}$ & AR \\
\noalign{\smallskip}
\hline
\noalign{\smallskip}
\hline
\noalign{\smallskip}
Hourglass~\cite{newell2016stacked} & 8\_Hourglass & 256$\times$192 & N & 25.1M & 14.3 & 66.9 & - & - & - & - & -\\
CPN~\cite{chen2018cascaded} & ResNet\_50 & 256$\times$192 & Y & 27.0M & 6.2 & 68.6& -& -& -& -& -\\
HRNet & HRNet\_w32 & 384$\times$288 & N & 28.5M & 16.0 & 69.0 & 84.7 & 75.8 & 66.2 & 77.4 & 79.0\\
CPN+OHKM  & ResNet\_50 & 256$\times$192 & Y & 27.0M & 6.2 & 69.4 &- &- &- &- &-\\
SimpleBaseline~\cite{xiao2018simple} & ResNet\_152 & 384$\times$288 & N & 68.6M & 35.5 & 70.2 & 88.2 & 77.3 & 66.8 & 77.1 & 76.1\\
SimpleBaseline & ResNet\_50 & 256$\times$192 & Y & 34.0M & 8.9 & 70.4& 88.6& 78.3& 67.1& 77.2& 76.3 \\

SimpleBaseline & ResNet\_101 & 256$\times$192 & Y & 53.0M & 27.9 & 71.4& 89.3& 79.3 &68.1& 78.1& 77.1\\
HRNet & HRNet\_w32 & 256$\times$192 & N & 28.5M & 7.1 & 72.1 & 89.5 & 78.6 & 69.5 & 78.0 & 78.6\\



SimpleBaseline & ResNet\_152 & 256$\times$192 & Y & 68.6M & 15.7 & 72.0 &89.3& 79.8& 68.7& 78.9& 77.8\\

HRNet~\cite{sun2019deep} & HRNet\_w32 & 256$\times$192 & Y & 28.5M & 7.1 & 74.4 & 90.5 & 81.9 & 70.8 & 81.0 & 79.8\\
HRNet & HRNet\_w32 & 384$\times$288 & Y & 28.5M & 16.0 & 75.8 & 90.6 & 82.7 & 71.9 & 82.8 & 81.0\\
HRNet & HRNet\_w48 & 384$\times$288 & Y & 63.6M & 32.9 & 76.3 & 90.8 & 82.9 & 72.3 & 83.4 & 81.2\\
\noalign{\smallskip}
\hline
\noalign{\smallskip}
1-stage & STNet & 256$\times$192 & N & 5.7M & 2.3 & 72.1 & 89.1 & 79.2 & 68.7 & 78.7 & 77.6\\
1-stage & STNet & 384$\times$288 & N & 5.7M & 5.2 & 73.2 & 89.1 & 80.0 & 69.4 & 80.1 & 78.6\\

2-stage & STNet & 256$\times$192 & N & 8.5M & 2.9 & 73.9 & 89.7 & 80.8 & 70.4 & 80.6 & 79.1\\
2-stage & STNet & 384$\times$288 & N & 8.5M & 6.5 & 75.6 & 90.0 & 81.8 & 71.8 & 82.5 & 80.6\\

3-stage & STNet & 256$\times$192 & N & 11.3M & 3.5 & 74.8 & 89.8 & 81.5 & 71.5 & 81.3 & 79.9\\
3-stage & STNet & 384$\times$288 & N & 11.3M & 7.8 & 76.2 & 90.4 & 82.4 & 72.5 & 82.9 & 81.2\\
3-stage* & STNet & 384$\times$288 & N & 20.3M & 12.6 & 76.8 & 90.7 & 83.3 & 73.3 & 83.4 & 81.8\\
\noalign{\smallskip} 
\hline
\end{tabular}
\end{center}
\end{table*}

\paragraph{Results on MPII} This dataset contains 25,000 images with 40,000 human instances labeled with 16 keypoints. The evaluation metric of the dataset is the \textit{Head-normalized Probability of Correct Keypoint (PCKh)} score. For MPII testing, we adopt the official testing strategy to use the provided human bounding boxes to estimate joints. We follow the six-scale testing procedure in~\cite{chu2017multi,tang2018deeply,yang2017learning}, and the PCKh@0.5 results are reported in Table~\ref{table:mpii}. As the parameters and layer numbers of three SimpleBaseline structures (ResNet\_50,101,152) increase gradually, their network capability becomes stronger. The results however show that stacking single receptive field layers tends overfitting without time-consuming pre-training. Table~\ref{table:mpii} shows the comparison results on MPII validation dataset with PCKh@0.5. As observed, STNet outperforms HRNet and SimpleBaseline with much fewer  parameters and GFLOPs. Compared to SimpleBaseline (ResNet\_152), the gains of three different STNet structures are 3.4\%, 4.3\% and 4.9\% with 92\%, 88\% and 84\% parameters drop.

\setlength{\tabcolsep}{0.5pt}
\begin{table*}[h!]
\tiny
\caption{Comparison results on the MPII validation set. }
\begin{center}
\label{table:mpii}
\begin{tabular}{lccccccccccccc}
\hline
\noalign{\smallskip}
Method & Backbone & Input Size & Pretrain & \#Param & GFLOPs & Head & Shoulder & Elbow & Wrist & Hip & Knee & Ankle & Mean \\
\noalign{\smallskip}
\hline
\noalign{\smallskip} 
\hline
\noalign{\smallskip} 
SimpleBaseline~\cite{xiao2018simple} & ResNet\_50 & 256$\times$256 & N & 34.0M & 12.0 & 96.5 & 94.8 & 87.7 & 82.2 & 87.9 & 82.3 & 77.1 & 87.6\\
SimpleBaseline & ResNet\_101 & 256$\times$256 & N & 53.0M & 16.5 & 96.1 & 94.3 & 86.1 & 80.3 & 87.2 & 81.4 & 76.2 & 86.6\\
SimpleBaseline & ResNet\_152 & 256$\times$256 & N & 68.6M & 21.0 & 95.7 & 93.7 & 84.9 & 77.9 & 85.5 & 79.2 & 74.3 & 85.2\\
HRNet~\cite{sun2019deep} & HRNet\_w32 & 256$\times$256 & N & 28.5M & 9.5 & 95.6 & 94.2 & 88.1 & 83.7 & 88.1 & 83.7 & 79.3 & 88.1\\
HRNet & HRNet\_w48 & 256$\times$256 & N & 63.6M & 19.5 & 96.0 & 94.6 & 88.8 & 84.1 & 87.4 & 83.6 & 80.4 & 88.4\\
TokenPose~\cite{li2021tokenpose} & L/D24 & 256$\times$256 & Y & 28.1M & - & 97.1 & 95.9 & 90.4 & 86.0 & 89.3 & 87.1 & 82.5 & 90.2\\
\noalign{\smallskip} 
\hline
\noalign{\smallskip} 
1-stage & STNet & 256$\times$256 & N & 5.7M & 3.1 & 96.7 & 94.9 & 89.0 & 83.6 & 88.4 & 84.4 & 79.7 & 88.6\\
2-stage & STNet & 256$\times$256 & N & 8.5M & 3.9 & 96.9 & 95.6 & 89.8 & 84.8 & 89.2 & 84.7 & 81.6 & 89.5\\
3-stage & STNet & 256$\times$256 & N & 11.3M & 4.6 & 97.0 & 95.9 & 90.2 & 85.2 & 89.6 & 86.4 & 83.1 & 90.1\\
\noalign{\smallskip} 
\hline
\end{tabular}
\end{center}
\end{table*}

\subsection {Ablation Analysis}
\begin{figure}[htbp]
  \centering
  \includegraphics[width=10cm]{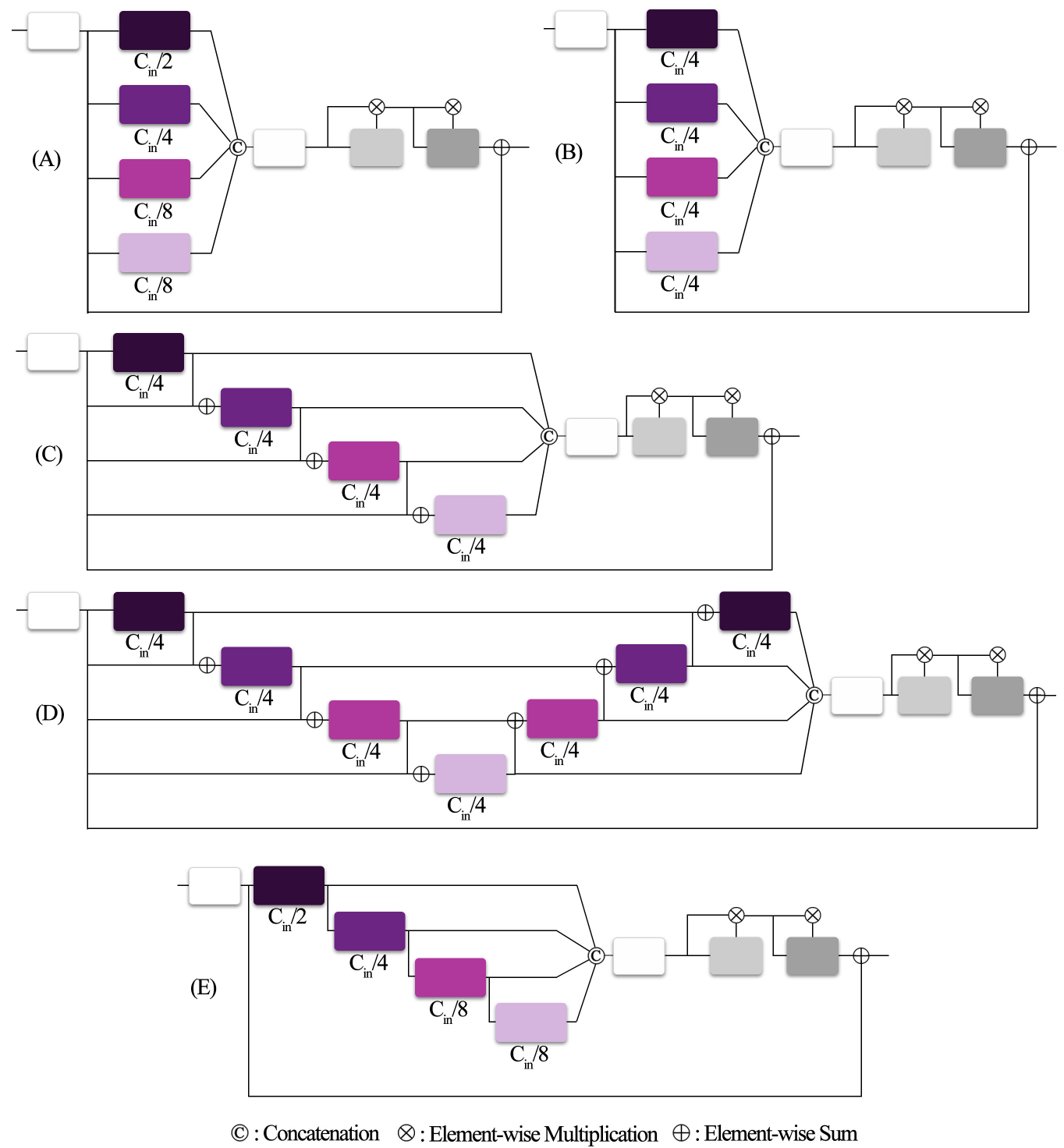}
  \caption{Different structures investigated in our paper. Structure E is adopted in this work.}
  \label{fig:STC_structures}
\end{figure}

\paragraph{STC Structure Exploration}
For the STC design, we explore a number of structures as illustrated in Fig.~\ref{fig:STC_structures}. Among multiple structures, though some  have fewer parameters and lower computational cost (Fig.~\ref{fig:STC_structures} (B), (C)), they lead to very limited performance practically. Meanwhile, we observe that channel separation manner is not preferable to extract significant local features. On the one hand, the channel separation manner leads to feature incoherence that the receptive fields of the following branch are not based on the previous one. On the other hand, the channel separation manner with the sum operation may cause some redundancies. By contrast, the final structure adopted by STC (Fig.~\ref{fig:STC_structures} (E)) attains the best performance with limited parameters and GFLOPs. 

\setlength{\tabcolsep}{3pt}
\begin{table}[htbp]
\scriptsize
\caption{Comparison results of multiple STC structures with 1-stage network. Structure E is adopted in this work.}
\begin{center}
\label{table:structure_compare}
\begin{tabular}{ccccccccc}
\hline
\noalign{\smallskip}
Structure & pretrain & Input Size & AP & AP$^{50}$ & AP$^{75}$ & AP$^{M}$ & AP$^{L}$ & AR\\
\noalign{\smallskip}
\hline
\noalign{\smallskip} 
\hline
\noalign{\smallskip} 
(A) & N & 256 $\times$ 192 & 71.5 & 88.9 & 78.5 & 68.0 & 78.3& 77.0\\
(B) & N & 256 $\times$ 192 & 70.6 & 88.7 & 77.9 & 67.0 & 77.5 & 76.3\\
(C) & N & 256 $\times$ 192 & 69.7 & 88.3 & 76.8 & 66.4 & 76.3 & 75.4\\
(D) & N & 256 $\times$ 192 & 71.5 & 88.7 & 78.8 & 67.8 & 78.4 & 77.0\\
(E) & N & 256 $\times$ 192 & \textbf{72.1} & \textbf{89.1} & \textbf{79.2} & \textbf{68.7} & \textbf{78.7} & \textbf{77.6}\\
\noalign{\smallskip} 
\hline
\end{tabular}
\end{center}
\end{table}

\begin{figure}[htbp]
  \centering
  \includegraphics[height=4.5cm]{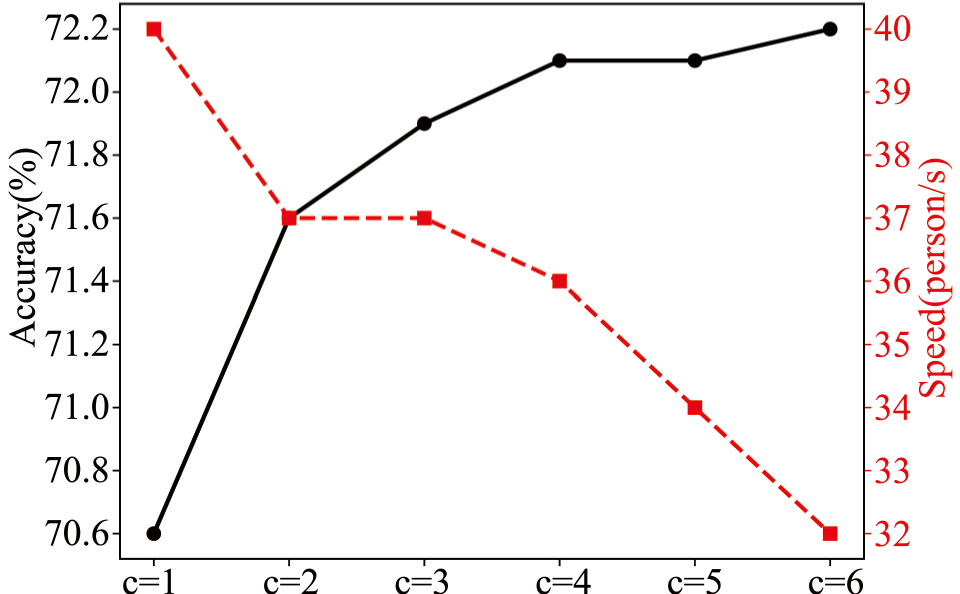}
  \caption{Accuracy and Speed of STC vs. branch numbers.}
  \label{fig:acc_fps}
\end{figure}

\paragraph{Ablation Analysis of STC} STC is proposed to focus on extracting multi-scale local representations through aggregating the outputs from four atrous convolution layers with different receptive fields. In this section, we take a closer comparison on STC to evaluate network performance with the different branch numbers. Table~\ref{table:STC_branch} shows the results on COCO validation dataset with the $256\times 192$ input size. As demonstrated in Table~\ref{table:STC_branch}, the performance gain is obviously in different stage networks, and the gain of accuracy becomes slight as we increase branch number more than 4. For 6-branch design, the channel number of the largest kernel size convolution is only 1 which brings a marginal increase. Fig.~\ref{fig:acc_fps} shows the 1-stage STNet inference speed of the network with different branches on COCO validation dataset where the inference speed gradually decreases without obvious accuracy improvement. As such, 4-branch design is adopted in this work.

\setlength{\tabcolsep}{2pt}
\begin{table}[htbp]
\scriptsize
\caption{Ablation analysis of STC with different branch numbers. $K$ means the kernel size.}
\begin{center}
\label{table:STC_branch}
\begin{tabular}{ccccccccccc}
\hline
\noalign{\smallskip}
\#Branch & Method & $K$ & AP & & AP$^{50}$ & AP$^{75}$ & AP$^{M}$ & AP$^{L}$ & AR\\
\noalign{\smallskip}
\hline
\noalign{\smallskip} 
\hline
\noalign{\smallskip} 
c = 1 & 1-stage & 3 & 70.6 & & 88.4 & 77.5 & 67.3 & 77.2 & 76.4 \\
c = 2 & 1-stage & 3,5 & 71.6 & & 89.1 & 78.8 &68.0 & 78.4 & 77.1 \\
c = 3 & 1-stage & 3,5,7 & 71.9 &  & 88.9 & 78.9 & 68.2 & 78.7 & 77.4 \\
c = 4 & 1-stage & 3,5,7,9 & 72.1 & 1.5 $\uparrow$ & 89.1 & 79.2 & 68.7 & 78.7 & 77.6 \\
c = 5 & 1-stage & 3,5,7,9,11 & 72.1 & & 89.1 & 79.1 & 68.7 & 78.7 & 77.5 \\
c = 6 & 1-stage & 3,5,7,9,11,13 & 72.2 & & 89.3 & 78.8 & 68.9 & 78.6 & 77.6 \\
\noalign{\smallskip} 
\hline
\noalign{\smallskip} 
\hline
\noalign{\smallskip}
c = 1 & 2-stage & 3 & 71.8 &  & 88.8 & 79.2 & 68.4 & 78.5 & 77.3 \\
c = 4 & 2-stage & 3,5,7,9 & 73.9 & 2.1 $\uparrow$ & 89.7 & 80.8 & 70.4 & 80.6 & 79.1 \\
\noalign{\smallskip} 
\hline
\noalign{\smallskip} 
\hline
\noalign{\smallskip}
c = 1 & 3-stage & 3 & 73.2 & & 89.3 & 80.3 & 70.0 & 79.8 & 78.6 \\
c = 4 & 3-stage & 3,5,7,9 & 74.8 & 1.6 $\uparrow$ & 89.8 & 81.5 & 71.5 & 81.3 & 79.9 \\
\noalign{\smallskip} 
\hline
\end{tabular}
\end{center}
\end{table}

\paragraph{Ablation Analysis of MDBS and STF} We now examine Multi-stage Dual Branch Structure, and STair Fusion on the COCO validation dataset. For simplicity, we take 1-stage and 2-stage STNets with the $256\times 192$ input size as the illustrative examples. Table~\ref{table:MDBS_STF} reports the performance when we gradually apply MDBS and STF on the networks. As observed, MDBS can lead to $0.6\%$ improvement than the baseline model on the 1-stage network. On the other hand, when STF is applied, compensation of the high-frequency information can further increase AP by $1.4\%$. Similarly, MDBS attains $1.7\%$ improvement and STF further achieves $0.6\%$ accuracy increase on the 2-stage network.
\begin{table}[htbp]
\scriptsize
\caption{Ablation analysis of STNet with different components. }
\begin{center}
\label{table:MDBS_STF}
\begin{tabular}{cccccccc}
\hline
\noalign{\smallskip}
Method & pretrain & Input Size & \#Params & GFLOPs & MDBS & STF & AP  \\
\noalign{\smallskip}
\hline
\noalign{\smallskip}
\hline
\noalign{\smallskip}
1-stage & N & 256$\times$192 & 3.35M & 1.74 & $\times$ & $\times$ & 70.1 \\
1-stage & N & 256$\times$192 & 3.37M & 1.79 & \checkmark & $\times$ & 70.7 \\
1-stage & N & 256$\times$192 & 5.74M & 2.32 & \checkmark & \checkmark & 72.1 \\
\noalign{\smallskip} 
\hline
\noalign{\smallskip}
2-stage & N & 256$\times$192 & 6.11M & 2.24 & $\times$ & $\times$ & 71.6 \\
2-stage & N & 256$\times$192 & 6.17M & 2.36 & \checkmark & $\times$ & 73.3 \\
2-stage & N & 256$\times$192 & 8.53M & 2.89 & \checkmark & \checkmark & 73.9 \\
\noalign{\smallskip} 
\hline
\end{tabular}
\end{center}
\end{table}

\paragraph{Ablation Analysis of STU Number} As the basic block of the multi-stage structure, STU can be simply stacked for adjusting the network capability. We perform comparison experiments on multiple architectures from one to three stages with two kinds of input sizes (256$\times$192 and 384$\times$288). We demonstrate the comparison results in Table~\ref{table:STU_block} where both the MDBS and STF mechanisms  are applied in this section, but pre-training process is not adopted.  Table~\ref{table:STU_block} demonstrates that the network performance is consistently improved with the increase of unit number. 

\setlength{\tabcolsep}{10pt}
\begin{table}[htbp]
\scriptsize
\caption{Performance of STU with different unit numbers.}
\begin{center}
\label{table:STU_block}
\begin{tabular}{ccccc}
\hline
\noalign{\smallskip}
Input Size & Pretrain & 1-stage & 2-stage & 3-stage \\
\noalign{\smallskip}
\hline
\noalign{\smallskip}
\hline
\noalign{\smallskip}
256$\times$192 & N & 72.1 & 73.9 & 74.8 \\
384$\times$288 & N & 73.2 & 75.6 & 76.2 \\
\noalign{\smallskip} 
\hline
\end{tabular}
\end{center}
\end{table}

\section{Visualization Analysis}
\begin{figure}[htbp]
    \centering
    \includegraphics[width=7cm]{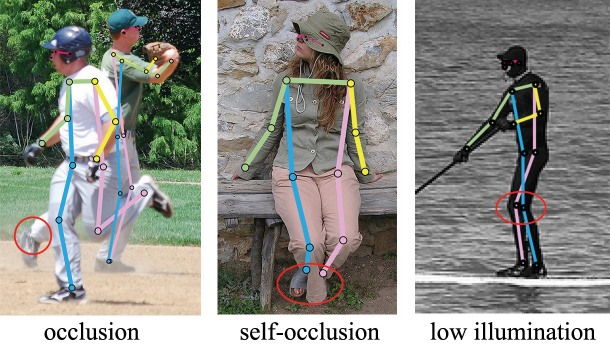}
    \caption{Several failure cases.}
    \label{fig:failure_case}
\end{figure}

\begin{figure*}[h!]
  \centering
  \includegraphics[width=12cm]{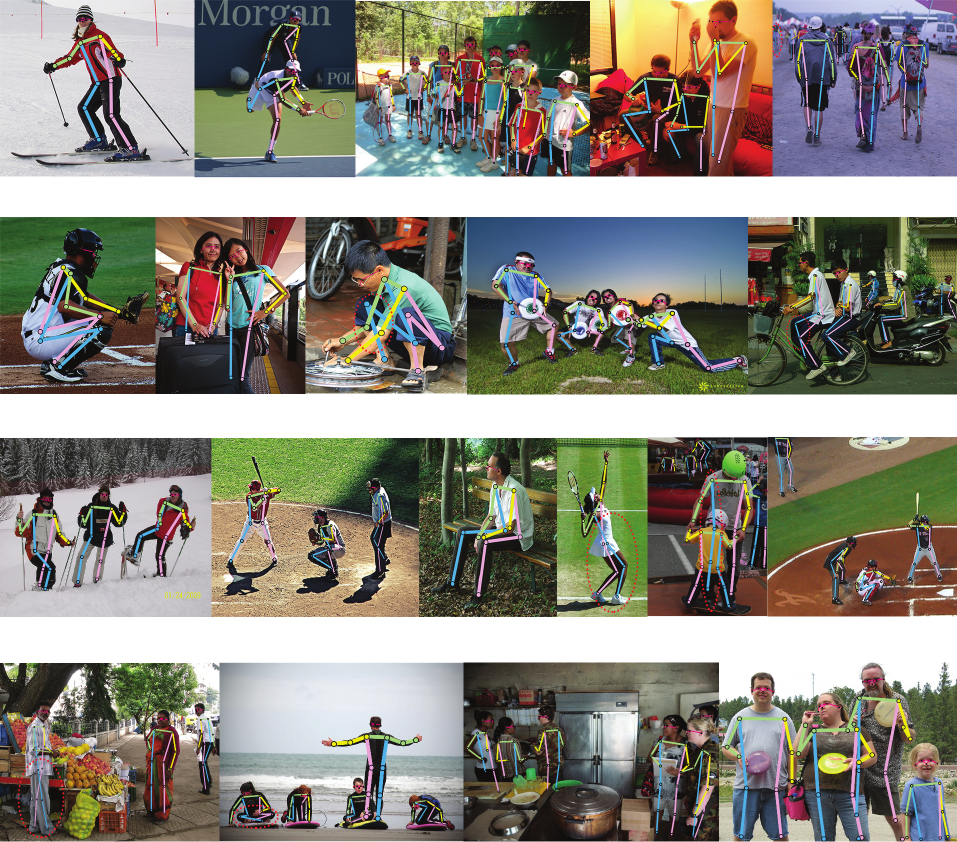}
  \caption{Successful cases visualization. The dotted circles denote some difficult scenes: small, vague, occlusion, ambiguity.}
  \label{fig:success_case}
\end{figure*}

Fig.~\ref{fig:success_case} visualizes some successful prediction cases, and the red dotted circles demonstrate some challenging scenes. The small and vague person subjects (5th case of 1st row), serious self-occlusion (3rd and 5th cases of 2nd row), serious occlusion (5th case of 3rd row) and ambiguous post (1st and 2nd cases of 4th row) can be predicted successfully due to the powerful local feature aggregation capability of STNet. In addition, Fig.~\ref{fig:failure_case} illustrates some failure cases of our method where the red circles point out the error predictions. The ambiguous occlusion and dark illumination are still challenges which we will focus in the future works.

\section{Conclusion}
In this work, we present a small yet effective multi-stage network for precise keypoint localization. To reduce the computational cost while maintaining superior performance, we propose a basic feature extraction block to focus on aggregating more diverse local representations through adopting multiple kernel sizes with fewer parameters. We alleviate the information loss problem from two aspects. Within each STair Unit, we keep high resolution feature maps to relieve feature loss. Outside the units, we develop a dual path structure to enhance feature re-usage and re-exploitation with low computational cost. Meanwhile, we design another mechanism to extract high-frequency texture representations. We test the effectiveness of our method through evaluations on standard pose estimation datasets, and the results demonstrate that the STNet's superiority with remarkable efficiency on parameters and GFLOPs.

\section*{Acknowledgements}
The work was partially supported by the following: National Natural Science Foundation of China under no.61876155; Natural Science Foundation of Jiangsu Province BE2020006-4; Key Program Special Fund in XJTLU under no. KSF-T-06, KSF-E-26.


\bibliography{arxiv}

\end{document}